\documentclass{article}

\usepackage{arxiv}

\usepackage[utf8]{inputenc} 
\usepackage{textcomp}
\usepackage[T1]{fontenc}    
\usepackage{hyperref}       
\usepackage{url}            
\usepackage[super,sort&compress,comma]{natbib} 
\usepackage{booktabs}       
\usepackage{amsfonts}       
\usepackage{nicefrac}       
\usepackage{microtype}      
\usepackage{lipsum}		
\usepackage{graphicx}
\usepackage{doi}
\usepackage{amsmath}
\usepackage{amssymb}
\usepackage{pifont}
\newcommand{\xmark}{\ding{53}}%

\title{Inverse deep learning methods and benchmarks for artificial electromagnetic material design}

 \author{ Simiao Ren \textit{$^{\ddag}$}, Ashwin Mahendra \textit{$^{\ddag}$},  Omar Khatib, Yang Deng,  Willie J. Padilla,and Jordan M. Malof \thanks{\textit{$^{\ddag}$} represent equal contribution. All other authors can be reached by firstname.lastname@duke.edu.} \\
	Department of Electrical and Computer Engineering\\
	Duke University\\
	Durham, NC 27708 \\
	\texttt{jordan.malof@duke.edu} \\
}

\renewcommand{\shorttitle}{\textit{arXiv} Template}

\hypersetup{
pdftitle={Inverse deep learning methods and benchmarks for artificial electromagnetic material design},
pdfsubject={q-bio.NC, q-bio.QM},
pdfauthor={David S.~Hippocampus, Elias D.~Striatum},
pdfkeywords={First keyword, Second keyword, More},
}

\begin{document}
\maketitle

\begin{abstract}
Deep learning (DL) inverse techniques have increased the speed of artificial electromagnetic material (AEM) design and improved the quality of resulting devices. Many DL inverse techniques have succeeded on a number of AEM design tasks, but to compare, contrast, and evaluate assorted techniques it is critical to clarify the underlying ill-posedness of inverse problems. Here we review state-of-the-art approaches and present a comprehensive survey of deep learning inverse methods and invertible and conditional invertible neural networks to AEM design. We produce easily accessible and rapidly implementable AEM design benchmarks, which offers a methodology to efficiently determine the DL technique best suited to solving different design challenges. Our methodology is guided by constraints on repeated simulation and an easily integrated metric, which we propose expresses the relative ill-posedness of any AEM design problem. We show that as the problem becomes increasingly ill-posed, the neural adjoint with boundary loss (NA) generates better solutions faster, regardless of simulation constraints. On simpler AEM design tasks, direct neural networks (NN) fare better when simulations are limited, while geometries predicted by mixture density networks (MDN) and conditional variational auto-encoders (VAE) can improve with continued sampling and re-simulation.
\end{abstract}


\section{Introduction}
In this work we consider the problem of designing artificial electromagnetic materials (AEMs), such as metamaterials, photonic crystals, and plasmonics.  The goal of AEM design is to find the geometric structure, material composition, or other features of an AEM - denoted $g$ - that will produce a desired electromagnetic (EM) response (e.g., a specific transmission or absorption spectrum), denoted $s$ here \cite{Chen2019,Staude2017,khatib2021deep}. This is a widely-studied problem involving a rich body of research, and a variety of effective methods \cite{herskovits1995advances}. Here we focus on emerging methods involving deep inverse models (DIMs), which have recently been found highly effective for solving AEM design problems \cite{khatib2021deep, huang2020deep,wiecha2021deep,jiang2020deep,ma2021deep}. 

DIMs are data-driven methods, and therefore assume access to a dataset of design-scattering pairs, $D=\{(g_n,s_n)\}_{n=1}^{N}$, which are obtained by evaluating the so-called "forward model" of the AEM system \cite{khatib2021deep} at specific values of $g$.  The output of the forward model, denoted $s=f(g)$, is usually estimated via theoretical results or computational electromagnetic simulations (CEMS).  DIMs then use $D$ to infer, or learn, an \textit{inverse} model, denoted $g=f^{-1}(s)$, that maps from a desired scattering directly to an AEM design that will produce the desired scattering.  This process of learning $f^{-1}$ is sometimes referred to as "training", and $D$ is referred to as the training dataset. 
Learning the inverse model is essentially a regression problem where $s$ comprises the independent variables, and $g$ comprises the dependent variable that we wish to predict.  Therefore conventional regression models, such as deep neural networks (DNNs), can be employed to infer $f^{-1}$.  A substantial body of recent work has investigated DNNs using deep neural networks (DNNs) to approximate the forward model, $f$, yielding impressive results \cite{khatib2021deep, Ma2020DeepStructures,So2020DeepNanophotonics,Mirosh2020, jiang2020deep}.  This success is often attributed to DNN's ability to approximate complex and highly non-linear functions.  Despite this capability, and its associated success however, DNNs can produce poor results if the problem is \textit{ill-posed} \cite{mueller2012linear, khatib2021deep}, as is often the case when approximating $f^{-1}$. 

\begin{figure}
 \centering
 \includegraphics[width=0.8\textwidth]{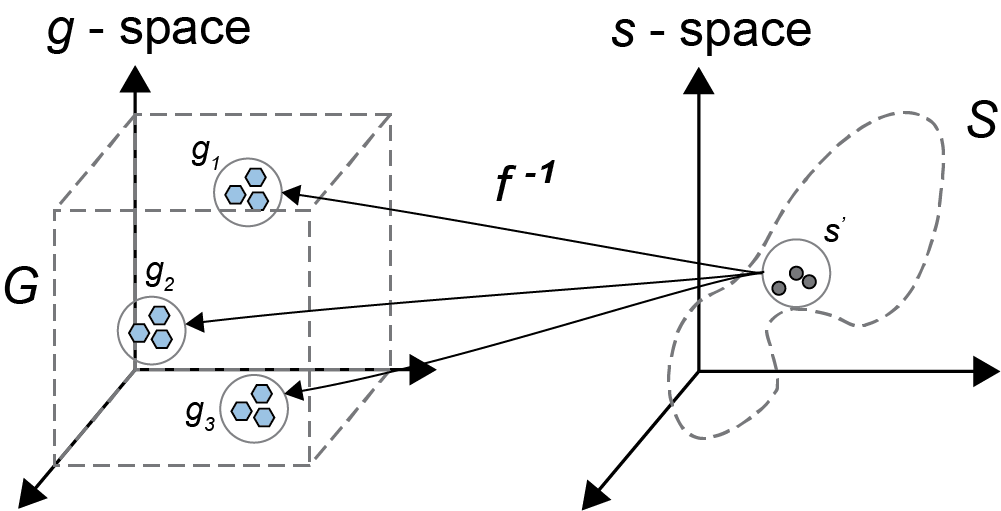}
  \caption{Illustration of non-uniqueness. There are multiple unique designs (left) that yield a similar scattering (right).  When trying to learn the inverse model (i.e., $g=f^{-1}(s)$), the input is $s$ and the model attempts to predict a valid value of $g$.  However, conventional regression models assume there is only one solution, and often fail when confronted with datasets where there are multiple valid solutions for some input values, as shown here. }
  \label{fig:non_uniqueness_concept}
\end{figure}

Ill-posed problems are those that violate one of the three following Hadamard conditions \cite{Hadamard1902}: (i) existence; (ii) uniqueness; and (iii) smoothness.  In principle, inverse AEM problems can violate any of the three Hadamard conditions, however, most recent attention has been given to violations of condition (ii), which is sometimes called "one-to-manyness", or non-uniqueness.  We refer the reader to other sources for a more thorough treatment of the other conditions \cite{khatib2021deep}. In the context of AEM design, non-uniqueness arises when there exist multiple values of $g$ that all yield similar AEM scattering (i.e., values of $s$), as illustrated in Fig. \ref{fig:non_uniqueness_concept}.  This is problematic for conventional DNNs because they can only produce a single output for each input, making it impossible to simultaneously predict two $g$ values for a given $s$.  Furthermore, the training procedure for DNNs does not account for non-uniqueness.  As a result, when DNNs are trained on datasets that exhibit non-uniqueness, they can learn to make highly inaccurate predictions.  


In recent years a large number of specially-designed models have been proposed, or adapted from the machine learning community, in order to overcome non-uniqueness in inverse AEM problems \cite{Liu2018GenerativeMetasurfaces,So2019DesigningNetworks,Ma2019ProbabilisticStrategy,Qiu2019DeepDesign, Liu2018TrainingStructures, Ma2018Deep-Learning-EnabledMetamaterials, ren2020benchmarking, deng2021,da2014optimization,Liu2020AStructures}. We refer to these approaches as DIMs, and they can be taxonomized into the following broad categories based upon their modeling strategy: probabilistic (GAN-based \cite{Liu2018GenerativeMetasurfaces,So2019DesigningNetworks},VAE-based \cite{Ma2019ProbabilisticStrategy,Qiu2019DeepDesign}), deterministic (e.g., tandem \cite{Liu2018TrainingStructures,Ma2018Deep-Learning-EnabledMetamaterials}), and iterative (e.g., neural-adjoint\cite{Peurifoy2018NanophotonicNetworks, ren2020benchmarking,deng2021}, genetic algorithms \cite{da2014optimization,Liu2020AStructures}).  Each of these these model classes has been employed successfully for solving AEM inverse problems.  We will describe each class of models in greater detail in Sec. \ref{sec:deep_inverse_model_descriptions}, as well as recent work employing them for inverse AEM problems.





\subsection{Challenges with benchmarking and evaluation}

Despite the rapid growth of DIMs in AEM research over the past several years, there has been little replication and benchmarking, making it difficult to determine the extent to which real methodological progress is being made over time.  This also makes it challenging for researchers and practitioners in the AEM community to choose methods that are best-suited to solve their problems.  One fundamental reason for the absence of benchmarking may be the difficulty of reproducing the computational simulations from a previous study, which requires substantial expertise and computation time.  Furthermore, publication often provide an insufficient level of detail to accurately reproduce simulations.  

One potential solution to this problem, which has become common in the machine learning community, is for authors to release their datasets and trained models. As noted in recent reviews \cite{khatib2021deep, jiang2020deep}, and quantified in a recent study  \cite{deng2021benchmarking}, this is rarely done in the AEM literature.  One notable exception to this trend is a recent study study that published a benchmark of three AEM problems for data-driven \textit{forward} modeling (i.e., approximating $f(g)$).  However, to our knowledge there has been no systematic comparison and analysis of DIMs for AEM problems.   

Another challenge with the study of DIMs is the selection of an appropriate inverse problem.  Few studies investigate whether the problem under consideration truly exhibits non-uniqueness, or compare to a conventional DNN.  These are important considerations since most DIMs are built upon the assumption that non-uniqueness is present in the data.  If there is no non-uniqueness, then any performance differences between inverse models must be due to other factors that are unrelated to ill-posedness.  Furthermore, simpler models such as conventional DNNs may achieve more accurate results in these cases. In this work we will show that not all inverse problems exhibit non-uniqueness, and in such cases it may be sufficient to simply use a conventional DNN.   

One final challenge with the study of DIMs, especially within the context of AEMs, is scoring.  One advantage of many modern DIMs is that they can propose multiple solutions for a given input (i.e., value of $s$).  Despite this capability, nearly all studies in AEM only evaluate the accuracy of the first proposed solution.  Furthermore, in practice it is often possible, or desirable, to evaluate the efficacy of several designs (e.g., via numerical simulation) and adopt the best one. Recent machine learning research has shown that modern DIMs can achieve substantially better results if several proposed solutions are considered, and that the best DIM depends on how many proposed solutions can be considered \cite{ren2020benchmarking}. In this work we will adopt a scoring metric that accounts for this capability.

\begin{figure*}[h]
\centering
  \includegraphics[width=0.8\textwidth]{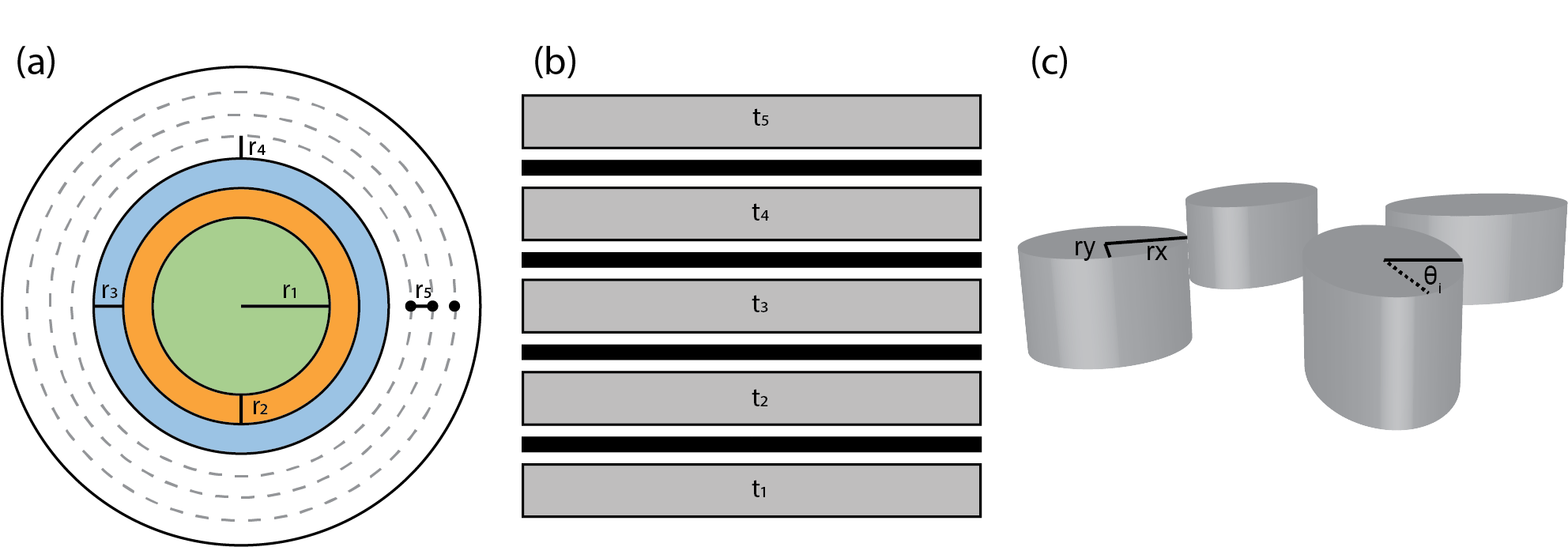}
  \caption{Depiction of the three artificial electromagnetic materials used for inverse design test. (a) concentric plasmonic layers composing the TiO\textsubscript{2}-Silica multi-layer Shell, (b) the multiple layers of alternating silicon nitride and graphene composing the Graphene-Si\textsubscript{3}N\textsubscript{4} 2D Multi-layer Stack, (c) 2$\times$2 unit-cell of all-dielectric metasurface resonators composing the .}
  \label{fig:materials_schematic}
\end{figure*}

\subsection{Contributions of this work}

\textit{(i) The first public and accessible benchmark of inverse AEM problems}.  In this work we develop a benchmark dataset comprising three unique inverse AEM problems: nanophotonic shell, graphene multi-layer stack, and an all-dielectric metamaterial array.  These problems were chosen to represent several unique sub-fields within AEM research, increasing the generalizability of any results obtained via the benchmark. Furthermore, we adopted our benchmark tasks from recently-published AEM research, helping to ensure their relevance and significance to the broader AEM community.  To support replication, we publish our benchmark datasets and documentation online.  Crucially, we also publish fast and easily-accessible forward simulators for each benchmark problem.  As we discuss in Sec. \ref{sec:benchmark_problems_and_design}, an accurate simulator is generally needed to evaluate the performance of DIMs, however making them fast and easily-accessible is not always trivial.       

\textit{(ii) The first systematic comparison of deep inverse models for AEM tasks.} Using our benchmark resources, we perform a systematic comparison of nine state-of-the-art DIMs.  One of these techniques, the conditional invertible neural network (cINN) is adapted from the machine learning community, and is novel in the context of AEM research. We systematically optimize each of these models and then compare their performance on the benchmark tasks.  Rather than evaluating their accuracy using a single proposed solution, following Ren et al.\cite{ren2020benchmarking} we evaluate how each model performs as a function of the number of proposed solutions that we are permitted to evaluate.  We assert that this is a more practically relevant measure of performance for inverse AEM tasks (see Sec. \ref{subsec:problem_formulation}). Importantly, we find that the relative performance of many DIMs depends upon the number of proposals that they are permitted to make. 

\textit{(iii) Identifying and addressing non-uniqueness in inverse AEM Problems.} Inverse problems are not guaranteed to exhibit non-uniqueness, which has several important implications when solving inverse AEM problems.  We discuss these implications and make several recommendations.  In particular, we suggest that researchers always include a conventional DNN as a baseline model when solving inverse problems, and we propose a measure for the level of non-uniqueness present in a given task. 



The remainder of this work is organized as follows: Section \ref{sec:benchmark_problems_and_design} introduces the design of our benchmark, including the benchmark problems, performance metrics, and data handling; Section \ref{sec:deep_inverse_model_descriptions} describes the eight DIMs in our benchmark comparison; Section \ref{sec:experimental_design} describes our experiments; Section \ref{sec:experimental_results} discusses the results of our experiment; and Section \ref{sec:conclusion} presents our final conclusions and future work.

\section{The inverse AEM benchmark} 
\label{sec:benchmark_problems_and_design}

The objective of our benchmark is to establish a shared set of problems on which the AEM community can compare DIMs, and thereby measure research progress more reliably.  To achieve this goal, we chose three initial problems to include in our benchmark, and we share resources to maximize the accessibility of the benchmark. 

\subsection{Problem formulation and error metrics}
\label{subsec:problem_formulation}

In data-driven inverse modeling we assume access to some dataset, $D=\{(g_n,s_n)\}_{n=1}^{N}$, comprising $N$ pairs of input and output sampled from the forward model of our system (i.e., AEM system in our case).  Each pair is generated by first sampling some input, $g \in G$, where $G$ refers to some designer-chosen domain.  In AEM problems $G$ is often a hypercube as illustrated in Fig. \ref{fig:non_uniqueness_concept}, and values are sampled uniformly \cite{deng2021, Chen2019, Peurifoy2018NanophotonicNetworks}.  We assume that $D$ is partitioned into three disjoint subsets: $D = D_{tr} \cup D_{val} \cup D_{te}$.  $D_{tr}$ is used to infer the parameters of the DIM (i.e., train the model), while $D_{val}$ is used to monitor the progress of training and stop it at an appropriate time (e.g., before over fitting).  $D_{te}$ is the testing dataset, and the goal of inverse modeling is to use $D_{tr}$ and $D_{val}$ to learn a model of the form $\hat{g} = \hat{f}^{-1}(s,z)$ that produces accurate designs for all $s \in D_{te}$. In other words, the designs inferred by $\hat{f}$ should yield scattering that is similar to the desired input scattering.  Many DIMs can produce multiple solutions for the same $s$, and the variable $z$ controls which of these solutions is output by the inverse model (see Sec. \ref{sec:deep_inverse_model_descriptions}).  

The error of DIMs is typically measured using re-simulation error, given by $\mathcal{L}(s,\hat{s}(z)))$, where $\mathcal{L}$ refers to some measure of the error between $s$ and $\hat{s}(z) = f(\hat{g}(z))$  (e.g., mean squared error (MSE)). We assume that we are permitted to evaluate $T$ proposed inverse solutions by passing them through the forward model (e.g., a computational simulator) so that we can compute their re-simulation error and then take the solution with the lowest error.  Then the objective of DIMs is to minimize the expected (i.e., average) error over the testing dataset, given $T$ solution proposals.  A sample estimator for this metric is given by  \cite{ren2020benchmarking}: 
\begin{equation}
    \hat{r}_T = \frac{1}{|D_{te}|} \sum_{s \in D_{te}} [\min_{i \in [1,T]} \mathcal{L}(\hat{s}(z_{i}),s)]  \label{eq:rt_estimate}
\end{equation}
where $Z_{i}$ is a set of $z$ values that is indexed by the variable $i$.  Note that, at $T=1$, the re-simulation error is equivalent to mean squared resimulation error, a widely-used metric in the AEM literature.  Therefore the goal of the DIMs will be to learn an inverse function that minimizes $\hat{r}_{T}$. 

For our benchmark, we propose to use MSE as the loss, since it is widely used for AEM problems \cite{Hou2020, li2019self,deng2021,Nadell2019,Kiarashinejad2020DeepNanostructures}.  It is also well-behaved and well-defined for all values of $s$, unlike the mean-relative-error -- another metric sometimes used in the AEM literature (e.g., Peurifoy et al. \cite{Peurifoy2018NanophotonicNetworks}, Chen et al. \cite{Chen2019}).  MRE has the limitation that it grows exponentially as the value of $s \rightarrow 0$, and becomes infinity when $s=0$.



\begin{figure}
 \centering
 \includegraphics[width=0.5\textwidth]{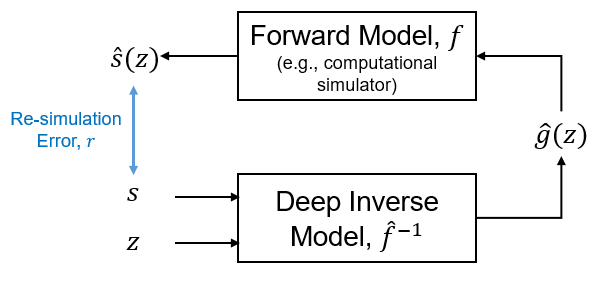}
  \caption{Illustration of re-simulation error. Note that the variable $z$ is only utilized by models that propose multiple solutions.}
  \label{fig:resimulation_error_concept}
\end{figure}

\subsection{Benchmark problems and selection criteria}
\label{subsec:benchmark_problem_descriptions}

The three AEM problems that we selected for inclusion in our benchmark are presented in Table \ref{Table:benchmark_data}, along with key details.  These problems were chosen based upon several criteria to maximize the relevance of our benchmark to the AEM community.  The first criterion was that each benchmark task had been studied in a recent AEM publication.  A second criterion was representativeness; we deliberately chose benchmark tasks that originate in different sub-fields within AEM research.  By choosing problems in this way we also help ensure that any conclusions obtained on the benchmark are more likely to generalize across AEM research.  Finally, we chose problems of varying input and output dimensionality, since dimensionality is an influential factor in the performance and behavior of DIMs.  Next we describe major details about each benchmark problem. 

\begin{table}[h]
    \begin{center}
    \begin{tabular}{cccc}
        \toprule
        Dataset name & Stack & Shell & ADM \\
        \midrule
        Description & 2D Multi-layer Stack & Multi-Layer Shell & All-Dielectric Metasurface \\
        Geometry Dimension, $|g|$  & 5 & 8 & 14\\
        Spectra Dimension, $|s|$ & 256 & 201 & 2,000 \\
        Training set size, $|D_{tr}|$ & 40,000 & 40,000 & 8,000 \\
        Validation set size, $|D_{val}|$ & 10,000 & 10,000 & 2,000 \\
        Test set size, $|D_{te}|$ & 500 &  500 &  500\\
        Source publication  & Chen et al.\cite{Chen2019} & Peurifoy et al.\cite{Peurifoy2018NanophotonicNetworks} & Deng et al. \cite{deng2021}\\
        \bottomrule
    \end{tabular}
\caption{Basic statistics about benchmark dataset used}
\label{Table:benchmark_data}
\end{center}
\end{table}

\textbf{TiO\textsubscript{2}-Silica Multi-layer Shell (Shell).} The geometry of a multi-layer dielectric spherical nanoparticle of alternating, tunable thickness TiO\textsubscript{2} and silica shells are optimized to produce target scattering-cross section spectra -- shown in Fig. \ref{fig:materials_schematic} (a). Peurifoy et al \cite{Peurifoy2018} describes this problem and implements an analytical Matlab simulator replicated in python for this work. Adjustable thickness TiO\textsubscript{2} and silica shells parameterize the geometry of the nanosphere. We consider the 8-parameter version of this problem, where spectra are discretized by 201 uniformly spaced outputs between the wavelengths of 400-800nm. 

\textbf{Graphene-Si\textsubscript{3}N\textsubscript{4} 2D Multi-layer Stack (Stack).} The geometry of a multi-layer stack of alternating graphene and Si\textsubscript{3}N\textsubscript{4} dielectric layers, as depicted in Fig. \ref{fig:materials_schematic} (b), is optimized to produce a target absorption spectra under an incident beam of s-polarized light. Chen et al.\cite{Chen2019} describes this problem and implements an analytical transfer matrix simulator. Paired graphene-Si\textsubscript{3}N\textsubscript{4} subunits of infinite width and adjustable Si\textsubscript{3}N\textsubscript{4} thickness parameterize the geometry of each stack. We consider the 5-parameter version of this problem, in which spectra are discretized by 256 uniformly spaced outputs between the wavelengths of 240-2000nm. 

\textbf{All-dielectric metasurface supercell (ADM).} This problem was originally described and published in \cite{deng2021}.  The ADM task was selected because it possesses several features: (1) It has 14-dimensional geometry inputs, as shown in Table \ref{Table:benchmark_data}, which is greater than many AEM studies found in the literature. The higher dimensional input results in greater complexity. (2) The scattering response in this dataset is the absorptivity spectrum with 2001 frequency points and many sharp peaks that are traditionally challenging to fit.  (3) This is the only dataset that is generated from full-wave simulation software. Each supercell, as shown in Fig \ref{fig:materials_schematic}, consists of four $SiC$ elliptical resonators. The geometry parameters of one supercell are: height $h$ (identical for all resonators), periodicity $p$, x-axis and y-axis radii $r_x, r_y$, and each elliptical resonator is free to rotate and described by $\theta$.  The absorptivity spectra are discretized by 1000 uniform frequency points from 100-500THz.   

\subsection{Neural surrogate simulators to enable replication.} One challenge with benchmarking DIMs is that it requires access to a computationally fast implementation of the true forward function, $f$; this is needed to compute the re-simulation error (Eq. \ref{eq:rt_estimate}) for each DIM.   Many AEM problems, such as the ADM problem here, rely upon computational simulators to evaluate $f$, which is time-consuming to setup, difficult to replicate across studies, and computationally slow. This is a major obstacle to evaluating DIMs on many modern AEM problems that rely upon computational simulation.  In this work we propose a general strategy overcome this problem, first suggested in Ren et al., \cite{ren2020benchmarking}, that involves training a data-driven surrogate model for the simulator (e.g., a deep neural network), $\hat{f}$, and then treating this surrogate as the true simulator for experimentation with DIMs.  A neural network surrogate is computationally fast that can be readily shared with other researchers for accurate replication and future benchmarking.  For the ADM problem here, we adopt a surrogate simulator that was developed in Ren et al. \cite{ren2020benchmarking}, which comprises an ensemble of neural networks trained on a large dataset from the original simulator. This surrogate model is computationally fast, and highly accurate ($6e-5$ with respect to the true simulator).  

\subsection{Benchmark resources.}
The (Python) code base for our benchmark task is maintained at the following remote repository: \url{https://github.com/BensonRen/AEM_DIM_Bench} and can be easily download. The code base includes all model architecture and code, and is open source under MIT license.  We will maintain our code base through a remote repository at github. This will allow users to post comments or concerns about the code, as well as build upon the code repository.  

\section{Invertible neural networks}
\label{sec:invertible_neural_network}

Invertible neural networks (INNs) and conditional INNs (cINNs) have recently been found to achieve state-of-the-art results for solving general data-driven inverse problems\cite{ardizzone2018analyzing,kruse2019benchmarking,ICML_invertible_workshop}, however they have yet to be explored for AEM problems. In this work we adapt INNs and cINNs to solve inverse AEM problems. We next summarize the technical details of these methods, however, readers can find a complete description for the INN in Ardizzone et al.\cite{ardizzone2018analyzing} and the cINN in Kruse et al. \cite{kruse2019benchmarking}.  

INNs are based upon flow-based models \cite{dinh2014nice, dinh2016density, kingma2018glow}, which assume that all data can be modeled as a sample from some non-elementary probability distribution, denoted as $S \sim p_S(s)$ where $p_S(s)$ is the probability distribution of S.  Then it is assumed that $S = f_{\theta}(Z)$, where $Z \sim p_{Z}(z)$ is a multi-dimensional Gaussian distribution with a diagonal covariance matrix (i.e., each random variable is independent), and $f$ is some function parameterized by $\theta$.  Although most real-world data (e.g., images or spectra) are not well-modeled by a multi-dimensional Gaussian, if $f_{\theta}$ is sufficiently complex then $p_S$ will also be sufficiently complex to accurately model real-world data.  

To achieve this level of complexity, the transformation is often implemented with a deep neural network, and the parameters $\theta$ are learned based upon real-world data.  Other contemporary models make the aforementioned assumptions (notably, the variational autoencoder) however, flow-based models (FBMs) are unique because they impose  structure on the deep neural network that ensures that it is bijective (a one-to-one mapping), and therefore invertible.  Furthermore, the inverse function $f^{-1}$ can be readily evaluated once the FBM is trained (i.e., once we infer $f$), so that it is trivial to evaluate by $s=f(z)$ or $z=f^{-1}(s)$

\textbf{Invertible Neural Network (INN).}
Inspired by the success of FBMs in modeling complicated distributions, \cite{dinh2016density, kingma2018glow} Ardizzone et al. \cite{ardizzone2018analyzing} adapted the FBM to solving inverse problems.  One constraint of the FBMs is that the dimensionality of its input and output must be identical, however, in most inverse problems the output is lower-dimensional than the input (i.e., $|s|<|g|$ using our notation).  To overcome this issue, they propose to formulate the forward model as $[s,z]=f(g)$, where $z$ is a normally distributed random variable.  The dimensionality of $z$ is chosen so that $|s|+|z|=|g|$. Then we train a FBM (i.e., an invertible neural network), $f_{\theta}$, that approximates the forward model, and where $\theta$ represents the network parameters.  Because the network is invertible, given a particular value of $s$ and a randomly-sampled value of $z$, we can also compute $\hat{g} = f(s,z)$. We can propose multiple solutions for a given value of $s$ by sampling multiple values of $z$ and passing them through $f^{-1}_{\theta}$ with the same value of $s$.  

The INN is trained by making forward passes (i.e., computing $[s,z]=f_{\theta}(g)$) for samples of data in the training dataset.  The parameters of the model are adapted to make accurate predictions of $s$ as well as making predictions of $z$ that are normally distributed. Mathematically, the objective function, or "loss", for training the INN is given by:  
 \begin{equation}
    \mathcal{L} = \frac{1}{2} \cdot \left( \frac{1}{\sigma^2} \cdot (\hat{s} - s)^2 + z^2\right) - \log|\det J_{g \mapsto[s,z]}|.\\ \label{INN:MLE}
\end{equation}
\begin{figure}[h!]
 \centering
 \includegraphics[width=0.8\textwidth]{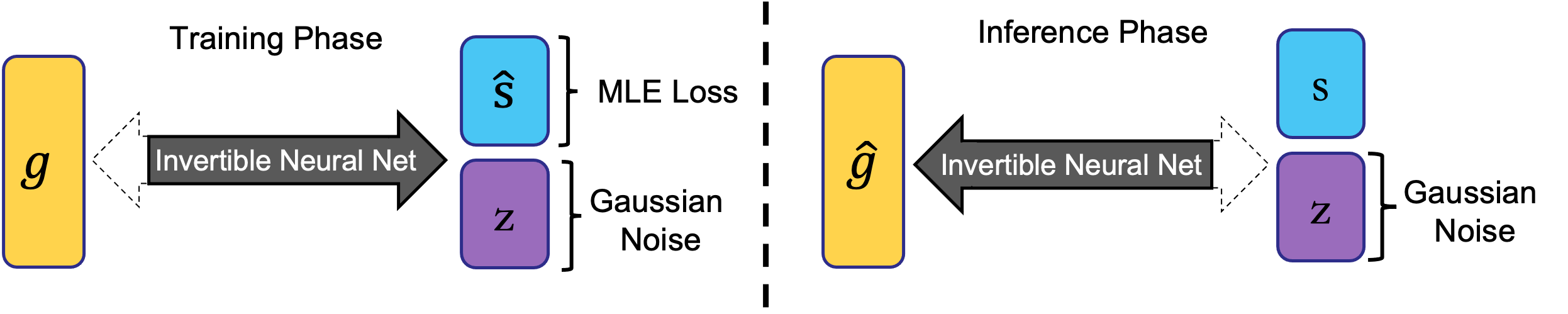}
 \caption{Schematic architecture of an Invertible Neural Network.}
 \label{arch_INN}
\end{figure}
Here $\sigma$ is a hyper-parameter controlling the trade-off between accurate $s$ prediction and the normality of $z$. The variable $J$ represents the Jacobian matrix of the mapping learned by the neural network to account for the probability volume change due to change of variable, which is trivially computed using invertible network structure . INNs can also optionally be trained with an  additional loss term comprising a maximum mean discrepancy (MMD) \citep{gretton2012kernel} measure, however, the INN was found to make more accurate predictions without it \citep{ardizzone2018analyzing}, and therefore we exclude it from our implementation. 

In all of the AEM problems considered in our benchmark the dimensionality of the output is substantially larger than the input (i.e., $|s|>>|g|$), breaking an assumption of the INN that $|s|<|g|$. To address this problem we follow the approach in \cite{ardizzone2018analyzing} and concatenate a vector of zeros to $g$ so that $|g|+|g_0| = |s|+|z|$, where $g_0$ is the vector of zeros.  This ad-hoc procedure is remedied by the cINN, which is described next.  


\textbf{Conditional Invertible Neural Network (cINN).}
The cINN was introduced in Kruse et al.\cite{kruse2019benchmarking}, and reformulates the inverse problem as $z = f(g|s)$, where $z$ is a vector sampled from a multi-dimensional Normal distribution.  Once again we use a neural network to model $f$ based upon data, however the goal is now to learn a mapping between $g$ and $z$ that is \textit{conditioned} upon $s$.  The authors propose an architectural change to the original invertible neural network so that the function can be conditioned on some auxiliary value (e.g., $s$ in our case) whether mapping in the forward or inverse direction.   This formulation implies that the dimensionality of $s$ and $g$ no longer need to match, and we simply need to set the dimensionality of $z$ to match $g$.  The cINN model trains with the following loss function:
\begin{equation}
    \mathcal{L} = \frac{1}{2} z^2 - \log|\det J_{g \mapsto z}|. \label{CINN:Loss}
\end{equation}
\begin{figure}[h!]
 \centering
 \includegraphics[width=0.8\textwidth]{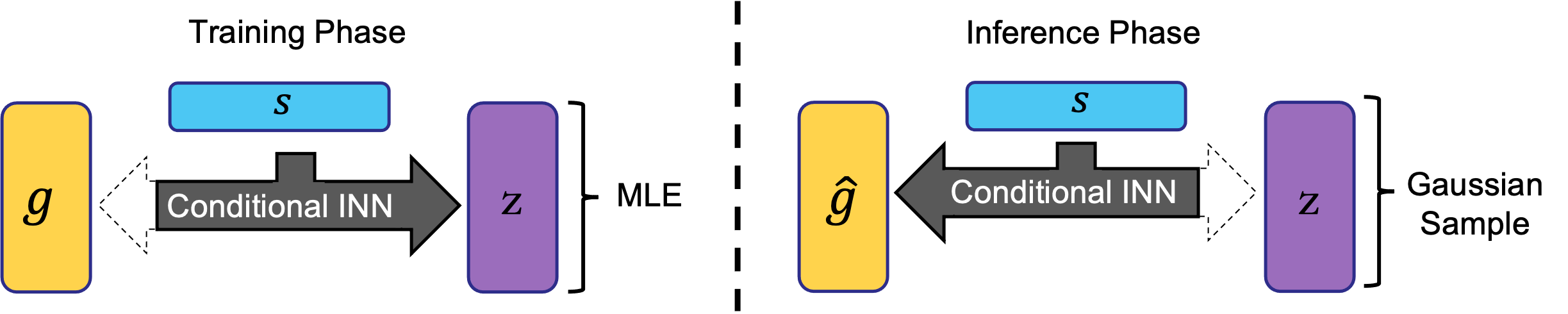}
 \caption{Schematic architecture of a conditional Invertible Neural Network.}
 \label{arch_cINN}
\end{figure}
Similar to the INN, the loss enforces the normality of $z$, however, $s$ no longer appears because the cINN is not tasked with predicting it. Once again $J$ represents the Jacobian matrix of the mapping learned by the neural network, this time when mapping from $g$ to $z$.



\section{Benchmark deep inverse models}
\label{sec:deep_inverse_model_descriptions}

In this section, we describe the DIMs that are included in our benchmarking experiments in Sec. \ref{sec:experimental_results}, as well as some key properties.  While we believe this section encompasses a large proportion of existing work on DIMs within the AEM community, it \textit{is not} intended to encompass all work.  We refer the reader to recent review articles for a more comprehensive treatment of the topic \cite{khatib2021deep}.  It is our hope however that AEM community will utilize our benchmark dataset and results to rigorously evaluate important additional models from the AEM literature in future work.

\subsection{Key inverse model properties for AEM applications}
\label{subsec:inverse_model_characterisctics}

Here we describe two key properties of DIMs with respect to AEM applications. Table \ref{Table:benchmark_inverse_model_characteristics} classifies each of our benchmark models according to several properties -- discussed next. The first important property is whether a DIM produces multiple solutions. As we find in Sec. \ref{sec:experimental_results}, models that are permitted to propose several solutions often also find progressively better solutions (i.e., $r_{T}$ reduces as $T$ grows). This capability also makes it possible for the designer to consider several viable solutions that may have somewhat different scattering, and choose the one that is best-suited for the application.  It is important to note that each additional model proposal that is considered requires an evaluation of the true forward model, $f$, which imposes a (usually modest) trade-off between design quality and computation time.  

A second important property of some DIMs is that they rely on an iterative process for inferring each inverse solution.  Most DIMs attempt to learn a direct mapping from $s$ to $g$, and therefore inference of a single solution proposal is computationally efficient.  In contrast, iterative methods usually make an initial set of guesses for the inverse solution, denoted $Z_{0}$.  A search for superior solutions is then performed based upon $Z_{0}$, and the results are used to update $Z_{0}$. This process is then repeated for some fixed number of iterations, or until the quality of the solutions (e.g., estimated resimulation error) no longer improves.  As we find in Sec. \ref{sec:experimental_results} iterative methods can often achieve the superior accuracy, although the computation required to infer solutions can be substantially larger than other methods. It is worth noting however, that this additional computation time is usually only a small fraction of the time for other processes, such as training the DIMs or evaluating $f$ using computational simulators.



\begin{table}[h]
    \centering
    \begin{tabular}{cccc}
        \toprule
        Model                                       & Multi-solution    &  Iterative    &  Applications to AEM problems  \\
        \midrule
        NN       & \xmark                &  \xmark           &   Chen et al. \cite{Chen2019}, Tahersima et al. \cite{Tahersima2019DeepSplitters}  Zhang et al. \cite{Zhang2019EfficientNetworks}            \\
        TD                                 &  \xmark             &   \xmark          &  Liu et al. \cite{Liu2018TrainingStructures}, Ma et al. \cite{Ma2018Deep-Learning-EnabledMetamaterials}, So et al. \cite{So2019SimultaneousNanoparticles}                  \\
        GA            & \checkmark               & \checkmark           &   Johnson et al.\cite{Johnson1997}, Zhang et al.\cite{Zhang2020}, Forestiere et al.\cite{Forestiere2012}                   \\
        NA                         & \checkmark               & \checkmark           &   Deng et al. \cite{deng2021}, Peurifoy et al. \cite{Peurifoy2018NanophotonicNetworks}, Ren et al. \cite{ren2020benchmarking}                    \\
        VAE             & \checkmark               &  \xmark           &    Ma et al. \cite{Ma2019ProbabilisticStrategy}, Qiu et al. \cite{Qiu2019DeepDesign} , Kudyshev et al. \cite{Kudyshev2020Machine-learning-assistedOptimization}               \\
        INN             & \checkmark               &  \xmark           &       -              \\
        cINN& \checkmark               &  \xmark           &       -              \\
        MDN               & \checkmark               &   \xmark          &       Unni et al. \cite{unni2021mixture, unni2020deep}                \\
        \bottomrule
    \end{tabular}
\caption{Summary of benchmark models}
\label{Table:benchmark_inverse_model_characteristics}
\end{table}

\subsection{Description of benchmark models}
\label{subsec:description_of_benchmark_models}
In this section we provide brief descriptions of each benchmark model.  We focus on describing the specific implementation that we use for our experiments, and its justification.  For a more thorough treatment of the models we refer readers to the references in Table \ref{Table:benchmark_inverse_model_characteristics}, or to recent reviews \cite{khatib2021deep}.  The major design parameters of each model were optimized according to the procedure described in Sec. \ref{subsec:model_optimization}, and further implementation details of each model can be found in the documentation of the code.


\textbf{Conventional Deep Neural Network (DNN).} In this approach the inverse problem is treated as a standard regression problem and a conventional DNN is used \cite{Chen2019, Tahersima2019DeepSplitters, Zhang2019EfficientNetworks, Akashi2020DesignMetamaterials, Lin2020cnnInverse}. Input is the spectra or response property of metamaterial and the output is the geometry properties. The backpropagation of a standard regression loss (e.g., mean-squared error) between the predicted and the true geometry updates the parameters of the neural network. No additional techniques are incorporated to compensate for one-to-many mappings between spectra and geometry. During training, the network is penalized for deviations between predicted geometries and the input spectrum's true geometry regardless of the re-simulation error between the target spectrum and spectrum of the predicted geometry. Residual connections, convolutional layers, batch normalization and dropouts can be added to boost performance when needed. In our specific implementation, we added batch normalization for all of three datasets for convergence acceleration.


\textbf{Genetic Algorithm (GA).} Genetic Algorithms (GAs), a set of iterative, optimization-based algorithms for inverse AEM design\cite{Johnson1997, Zhang2020,Forestiere2012}.  Our GA closely follows the model employed in Zhang et al\cite{Zhang2020} and Forestiere et al\cite{Forestiere2012} to solve inverse AEM problems. The GA first produces a set, or population, of initial geometries by randomly sampling each parameter of the geometry from a uniform distribution. Each geometry is then passed through the forward model of the process, $s=f(g)$, so that its re-simulation error can be computed with respect to the target spectrum.  This re-simulation error is used to determine the quality or "fitness" of each geometry in the population.  The traditional implementation of the GA relies on evaluating the true forward model for each candidate geometry, which can be computationally intensive if this requires computational simulation, as is the case for our ADM problem. To overcome this obstacle, we train a deep neural network to approximate $f$, and use this to compute the re-simulation error, dramatically accelerating the inference time of the GA. 

Once the fitness of each candidate geometry is comptued, the next step is the selection of “parent” geometries from this population. Crossover and mutation operations on this “parent” subset produce the next generation. We utilize a “roulette-wheel” selection pattern, as implemented by Zhang et al\cite{Zhang2020} and Forestiere et al\cite{Forestiere2012}, to select parent geometries. In roulette-wheel selection, the probability of a particular geometry being selected is proportional to that geometry’s fitness. Our GA also implements “elitism”, whereby copies of a certain number (determined by the elitism hyper-parameter) of geometries of absolute greatest fitness automatically pass to the future child population without undergoing mutation or crossover.

\textbf{Neural Adjoint (NA).} Neural Adjoint (NA) method was recently proposed in Ren et al. \cite{ren2020benchmarking} and subsequently found to be highly effective for solving AEM inverse problems Deng et al. \cite{deng2021}.  The NA is a gradient-based optimization approach that was developed based upon similar earlier methods employed in the AEM community \cite{Peurifoy2018NanophotonicNetworks, Asano2018OptimizationLearning, miyatake2020computational, deng2021, ren2020benchmarking}. The NA relies on the pre-trained forward network that maps from $g$ to $s$ space accurately as a proxy forward model to provide analytical gradient feedback during inverse inference. After the training of the forward network, by freezing the network weights and setting the input geometries as trainable parameters, it iteratively trains a set of (randomly) initial geometries to approach the target spectrum output by backpropagation of a mean squared error and boundary loss over the output and target spectra. The key difference between the NA and prior approaches is an additional boundary loss term \cite{ren2020benchmarking}, which steeply penalizes geometries outside the training domain of the proxy forward neural network to ensure that the network only returns $g$ values that are within the domain of the training data (where the proxy model makes accurate predictions). The boundary loss was found to greatly improve the performance compared to prior methods, and therefore we adopt the NA from Ren et al. \cite{ren2020benchmarking} here.  

\textbf{Tandem Network (TD).} The so-called data collision problem causes unstable gradient signals, and one method proposed to mitigate this issue is the tandem structure\cite{ Liu2018TrainingStructures,Ma2018Deep-Learning-EnabledMetamaterials,Ashalley2020MultitaskMetamaterials,So2019SimultaneousNanoparticles,Malkiel2018PlasmonicLearning,Pilozzi2018MachinePhotonics,Long2019InverseLearning,Xu2020EnhancedApproach,Hou2020PredictionLearning,He2019PlasmonicNanoparticle, Gao2019BidirectionalSilicon, Mall2020FastDesign,Phan2020GraphenePlasmons, Singh2020cCNN}.  TDs first train a neural network, the forward network, capable of accurately solving the forward problem from $g$ to $s$ with standard regression loss (MSE). The forward network’s weights are then fixed and its inputs are tied to the outputs of a second, separate network. The combined network is trained on the inverse problem. The fixed-weight forward network approximates a geometry-to-spectra simulator, training the pre-pended network with a standard regression loss between the predicted and input spectrum. The pre-pended network then learns a one-to-one relationship from input spectra and output geometries. As there is not much variation between TD architectures, we used the initial implementation of \cite{Liu2018TrainingStructures} with the modification to add a boundary loss term that was proposed in Ren et al. \cite{ren2020benchmarking}, since this was found to significantly improve its performance. 


\textbf{Mixture Density Network (MDN).} MDNs\cite{Bishop1994} account for the one-to-many nature of the inverse problem by assuming the distribution of conditional probability $p(g|s)$ is a mixture of Gaussian random variables. Its effectiveness has been shown on recent AEM design \cite{unni2020deep, unni2021mixture, ren2020benchmarking}. The number of Gaussian mixtures approximates the number of "many" in the one-to-many setting (hardly known beforehand) and is a sensitive hyper-parameter that has to be tuned carefully using a validaton set. We used the same implementation of the MDN as that presented in the literature where a diagonal covariance matrix were used. \cite{unni2020deep}


\textbf{Conditional Variational Autoencoder (VAE).} VAEs \cite{Kingma2014} also model the inverse problem probabilisticly and have shown successful design results in AEM\cite{Ma2019ProbabilisticStrategy,Ma2020AStructures, Kudyshev2020Machine-learning-assistedOptimization,kudyshev2020machine,Qiu2019DeepDesign,shi2020metasurface,Liu2020AStructures, Kiarashinejad2020DeepNanostructures}. 
Similar to an auto-encoder, where the inputs are encoded into latent space and decoded back for the reconstruction of the original input signal, VAEs uses a variational approach and encourage the latent space to become a Gaussian distribution. In AEM design space, the encoder  ($p(z | g, s)$) encodes $g$ into latent space $z$ conditioned on $s$, and the decoder ($p(g | z, s)$) decodes the sampled latent space variable $z$ into $g$ conditioned on $s$ as well. During the inference phase, a random sample is drawn from the latent space and passed to the decoder to get the inverse solution conditioned on a given desired $s$. We used the original architecture from Ma et al. \cite{Ma2020AStructures} without the convolution layers as none of our geometries are 2D and a parametric assumption of Gaussian latent space is effectively the same but architecturally simpler than the adversarial approach proposed by Kudyshev et al. \cite{Kudyshev2020Machine-learning-assistedOptimization}.


\begin{figure*}[h!]
    \begin{center}
        \includegraphics[width=\textwidth]{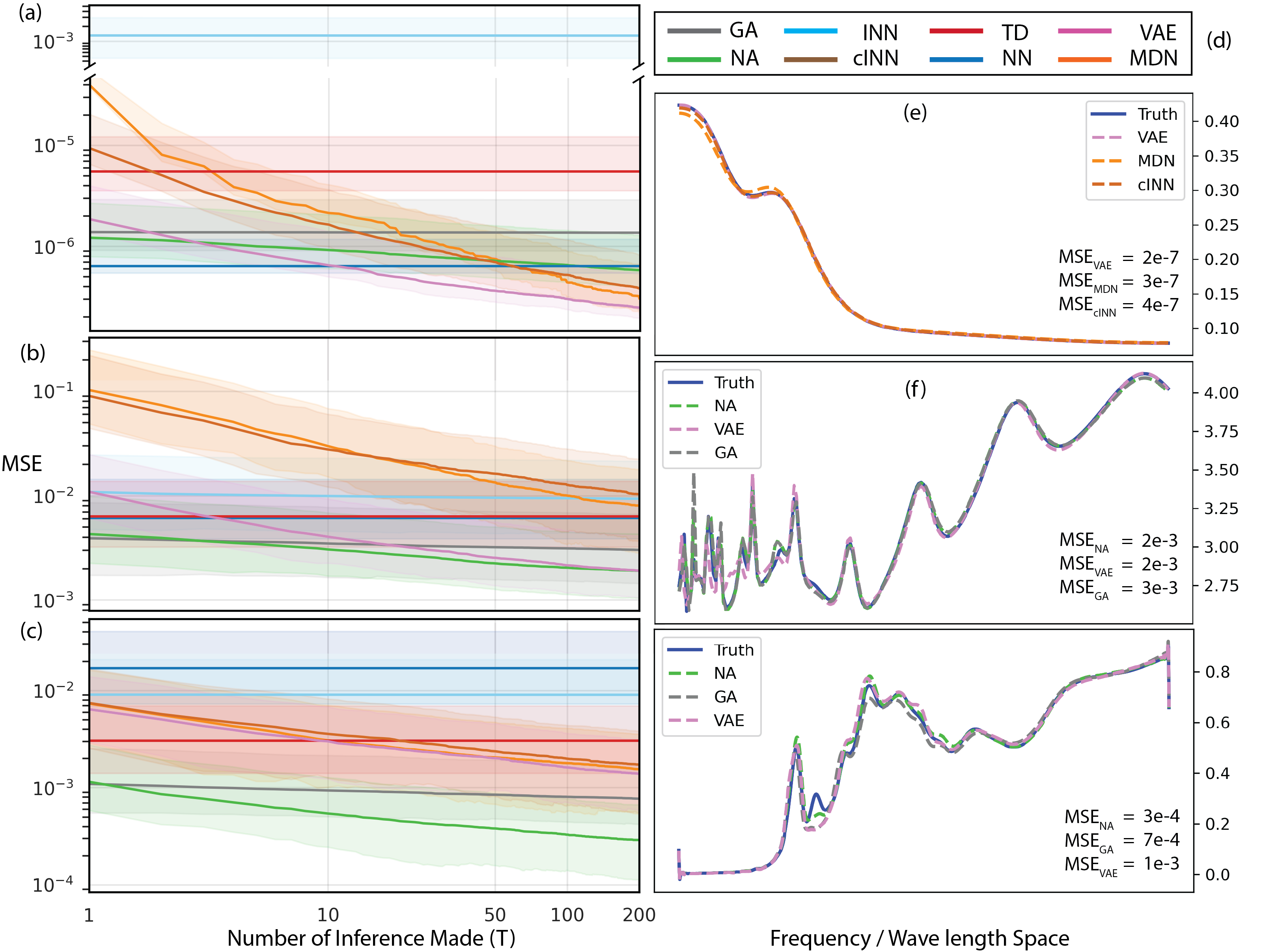} 
         \caption{Resimulation error for nine different inverse approaches for three benchmark data sets as $T$ increases, (a) Graphene-Si\textsubscript{3}N\textsubscript{4} 2D Multi-layer Stack, (b) TiO\textsubscript{2}-Silica Multi-layer Shell, and (c) All-Dielectric Super-unit Cell. Highlighted regions encapsulate 75$^{th}$ and 25$^{th}$ percentiles of resimulation error. $T=1$ performance ranked on left vertical axis, $T=200$ performance ranked on right vertical axis. (d) Legend of the plot (a-c). (e-f) The example spectra plot for T=200 for the top-3 methods shown from figure (a-c). Each spectra plotted is approximately (differ by less than 10\%) the MSE of that T=200 performance, labelled using text box at right bottom corner. The ground truth target spectra is plotted with solid blue line.}
         \label{Fig:resimulation_error_versus_T}
     \end{center}
\end{figure*}

\section{Experimental design} \label{sec:experimental_design}

The fundamental goal of this work is to establish a rigorous comparison of the above-mentioned deep inverse models applied to different problems in metamaterial design. The overall design of our experiment is as follows: (1) Collect data: we generate the training data using code as section \ref{sec:benchmark_problems_and_design} illustrated; (2) Train the model as described in section \ref{sec:deep_inverse_model_descriptions} on the training set; (3) Evaluate each model's ability to reproduce the target spectra $s_{gt}$ in the test set for every dataset, through re-simulation of the 200 predicted candidate geometry solutions $\hat{g}$.

\subsection{Model optimization and training}
\label{subsec:model_optimization}

For each dataset-DIM pair, we use 24 GPU-hours (specification of hardware used can be found in the appendix Section 6) for network training and hyper-parameter optimization (e.g., width and depth of models, learning rates, regularization strength, model-specific parameters like number of Gaussian for MDN and VAE, KL-divergence coefficient for VAE, MSE coefficient and padding dimension for INN and cINN, mutation rate of the GA etc.. We used a uniform grid of several hyper-parameters dimensions, adjusting the grid range and granularity from heuristics). This was sufficient to evaluate more than 80 hyperparameters settings for each model-dataset pair.  We empirically found this number sufficient to achieve diminishing performance returns for each model.  During this optimization process, we trained all models on the same training dataset, $D_{tr}$, for each problem.  All model (all implemented in Pytorch \cite{Pytorch2019}) are trained using Adam \cite{kingma2014adam} with batchnorm layer and a learning rate scheduler that reduces on plateau of training loss. We used batch size of 1024 and 300 epochs (by which time all models reached convergence).   

To evaluate the quality of a particular hyperparameter setting during the optimization process we measured $r_{T=1}$ on $D_{val}$ and ultimately chose the model for final testing that achieved the lowest validation error.  The final estimate of error for each model-task pair was evaluated by computing $r_{T}$ on $D_{te}$.  The final size and run time of each model-dataset pair can be found in the appendix.

\begin{table}[h]
\begin{center}
  \begin{tabular}{c|cc|cc|cccc}
    \toprule
    \multicolumn{1}{c|}{}& 
    \multicolumn{2}{c|}{Iterative}&
    \multicolumn{2}{c|}{Deterministic}&
    \multicolumn{4}{c}{probabilistic} \\
    \midrule
    T=1 & NA & GA  &  NN & TD & INN & cINN & MDN  & VAE\\
    \midrule
    Stack  & 1.22e-6 & 1.39e-6 & \textbf{6.37e-7} & 4.37e-6 & 1.30e-3 & 9.38e-6 & 3.95e-5 & 1.86e-6 \\
    Shell  & 3.60e-3 & \textbf{3.91e-3} & 6.12e-3 & 7.13e-3 &  1.08e-2 & 9.03e-2 & 1.02e-1 & 1.09e-2\\
    ADM  &  1.16e-3 & \textbf{1.10e-3} & 1.72e-2 &  1.66e-3 & 9.08e-3 & 7.45e-3 & 7.34e-3 & 6.42e-3\\
    \bottomrule
    T=200 & NA & GA  &  NN & TD & INN & cINN & MDN  & VAE\\
    \midrule
    Stack  & 5.83e-7  & 1.36e-6 & - & - & 1.29e-3 & 3.78e-7 & 3.03e-7  & \textbf{2.46e-7}  \\
    Shell  &  \textbf{1.58e-3} & 3.02e-3 & - & - &  9.36e-3 & 1.03e-2 & 8.05e-3 &  1.91e-3 \\
    ADM  &  \textbf{3.00e-4} & 7.73e-4 & - &  - & 9.05e-3 & 1.73e-3 & 1.55e-3 & 1.39e-3 \\
    \bottomrule
    
  \end{tabular}
  \caption{Table of $T=1$ re-simulation error over different DL inverse techniques on each benchmarking task. The best performing methods are in bold. For the deterministic ones, since they do not have the ability to generate multiple solutions, the T=200 accuracy is the same as T=1 and therefore omitted.}
  \label{Table:tabulated_resimulation_errors_for_T1_and_T200}
  \end{center}
\end{table}

\section{Results and discussion} \label{sec:experimental_results}


Our experimental results are presented in Fig. \ref{Fig:resimulation_error_versus_T}(a-c), where we plot re-simulation error ($r_{T}$, for $T\in[1,200]$) for each of the three benchmark tasks.  The results indicate that an error of approximately $10^{-3}$ (or much lower) can be achieved for all of the benchmark tasks. The achievable error does vary significantly across the three tasks however; for example, the DIMs reach $10^{-7}$ on the Stack problem, and only $\sim 10^{-3}$ for the Shell problem.   In Fig. \ref{Fig:resimulation_error_versus_T}(e-g) we plot a sample target spectrum (solid line) and the solutions produced by the top-three-performing models (dashed lines) for each task. These visualizations illustrate the high complexity of the target spectra, as well as the levels of accuracy that can be achieved by the DIMs.  We believe that these levels of error are sufficient to support a variety of different AEM research and development applications, and therefore demonstrate the overall effectiveness of DIMs across a variety of AEM problems. 

A major objective of these experiments is to compare the performance of state-of-the-art DIMs. This is challenging because the relative performance of the DIMs depends strongly upon the number of permissible solution proposals (i.e., value of $T$).  This is because the error of multi-solution DIMs (see Table \ref{Table:benchmark_inverse_model_characteristics}) often reduces significantly as $T$ increases, whereas single-solution models do not. The results therefore suggest that multi-solution DIMs offer substantial value if multiple solutions can be considered during design.  Furthermore, as we will show, the best-performing model and its achievable performance depend strongly upon $T$. This is an important finding because most existing AEM research only measures the performance of inverse models for $T=1$, and no existing studies evaluate $T>1$ in the fashion we have here.  Since it often requires thousands of simulations to train DIMs, we assert that it should usually be possible to consider many solutions, i.e., $T>>1$, making this an important performance measure. 

Given the dependency of performance on $T$, it is difficult to make general statements about the best-performing DIMs. To simplify this analysis we discuss the performance of the DIMs under two more specific scenarios that are especially relevant to AEM applications: when $T=1$ and $T \rightarrow \infty$, respectively.  In the last section we discuss the computation time required for each of the DIMs, and its implications for selecting an appropriate DIM. 

\subsection{Performance when $T=1$}
\label{subsec:performance_T_1}
The vast majority of existing AEM literature evaluates DIMs using $r_{T=1}$ (i.e., mean-squared re-simulation error) making it an important scenario to consider in our benchmark.  In Table \ref{Table:tabulated_resimulation_errors_for_T1_and_T200} we present $r_{T=1}$ for each DIM on our three benchmark tasks. Overall, the NA and GA methods achieve the lowest error, and achieve similar results on all three tasks.  They achieve the lowest error on the Shell and ADM tasks, however they are outperformed by the NN model on the Stack task. In Sec. \ref{sec:identifying_and_addressing_non_uniqueness} we show that the Shell problem does not exhibit significant non-uniqueness, making it suitable for conventional regression models, and making it disadvantageous to use DIMs. 

Given these results, the best-performing DIMs on our benchmark are iterative (see Table \ref{Table:benchmark_inverse_model_characteristics}), and their performance advantages come at the cost of somewhat greater computation time compared to other DIMs.  However, we note that the additional computation time is small for the tasks here: e.g., 60 seconds, or less (see Sec. \ref{subsec:computational_costs}).


\subsection{Performance when $T \rightarrow \infty$.}
In this subsection we consider the relative performance of DIMs when they are permitted to make a large number of solution proposals.  It is computationally intensive to evaluate their asymptotic performance (as $T \rightarrow \infty$) for all experiments, and therefore we use $T=200$ as an approximation.  

We argue that this is an important performance measure for inverse AEM problems because it closely reflects the goals of AEM researchers in practice.  Most often researchers want the best-performing design, and will evaluate numerous candidate designs in pursuit of this goal.  When using DIMs in particular, the quantity of data needed to train the models is substantially larger than $T=200$, implying that evaluating $T=200$ solutions (or more) with a simulator will often be feasible.  Furthermore, and as we show here, doing so often yields substantially-better designs. Therefore we believe this performance measure is of particular interest to the AEM community.  To our knowledge we are the first to propose this measure for evaluating the performance of inverse solvers on AEM problems. 

In Table \ref{Table:tabulated_resimulation_errors_for_T1_and_T200} we also present $r_{T=200}$ for each DIM on our three benchmark tasks. As expected, the error of most multi-solution DIMs reduces substantially as more solution proposals are permitted.  In this scenario the NA method achieves the lowest error on the Shell and ADM problems. The next best-performing models are the GA and the MDN.     

Regarding the Stack task, all of the models achieve very low error rates ($<10^{-5}$).  As discussed later in Sec. \ref{subsec:performance_T_1}, the Stack problem does not exhibit significant non-uniqueness, suggesting that superior performance does not imply superior ability to address non-uniqueness.  In this case many DIMs still achieve reductions in error as they make more proposals because they are making slight improvements in their estimates of the \textit{same} inverse solution, rather than a superior unique solution.  As a result, the improvements in accuracy are very small as $T$ increases (note the logarithmic error scale).

\subsection{Model computation times}
\label{subsec:computational_costs}

Another consideration when utilizing DIMs is their computation time: specifically their training and design-inference time.  The precise computation time of a DIM will depend strongly upon the particular task, model size (number of free parameters), and hardware being utilized; a full discussion of these dynamics is beyond the scope of this work. Therefore we present benchmark timings of each DIM on our three benchmark tasks, using a common hardware configuration (a single Nvidia RTX 3090 graphics processing unit). These measures therefore provide rough estimates of computation time that can be expected in many real-world settings for each model. The results of our benchmark timings are provided in Table \ref{Table:train_test_time}.  

In general the MDN, cINN, and INN models require the longest time to train, while the NN, VAE, GA and NA tend to require the least amount of time. The training times also vary across tasks, with the ADM task requiring substantially less time than the others.  This is likely due to the smaller quantity of training data available compared to the other problems.  Regarding inference time, the NA and GA are substantially slower than the other models, due to their iterative inference procedure.  These models tend to achieve the best performance and therefore this imposes a trade-off between design quality and computation time.  However, we note that the inference time of the models is relatively small compared to their training time.  Furthermore, the time required for computational simulations is usually (though not always) much greater than the time required for training and (especially) solution inference with DIMs.  Ultimately the relevance of these factors, and the most appropriate trade-offs, will be  problem dependent.  

\begin{table}[h!]
    \begin{center}
    \begin{tabular}{c|cc|cc|cc}
        \toprule
        Model     & 
        \multicolumn{2}{c|}{Stack} &  
        \multicolumn{2}{c|}{Shell}   & 
        \multicolumn{2}{c}{ADM}  \\
          & Train & Eval & Train & Eval & Train & Eval \\
        \midrule
        NN       & 655 & 0.0035 &  590 & 0.0030 & 50  &   0.0036 \\
        TD       & 428&  0.0029  &   856 & 0.0027& 198 & 0.0030       \\
        GA       & 202 &   24 &  519& 26  &  72  & 25\\
        NA       & 202  & 1.4&  519 & 6.65 &  72 & 5.21\\
        VAE     & 320  & 0.0023 &  363 & 0.0026 &  50 & 0.0027 \\
        INN      & 1843 & 0.0079 &  2919 & 0.015 &  487 & 0.0124\\
        cINN     & 541 & 0.013 &  1131 & 0.019 &  152 & 0.0096\\
        MDN      & 350 &  0.015 &  434 & 0.003&  144 & 0.010 \\
        \bottomrule
    \end{tabular}
\caption{Model training and evaluation time (unit of seconds) of each individual models. Note that training time is dependent of the dataset size and is nearly a constant given hyper-parameter, dataset size and same computation resources (in appendix). The evaluation time is averaged over getting 200 solutions (proposals) without taking IO time into account.}
\label{Table:train_test_time}
\end{center}
\end{table}

\section{Identifying and addressing non-uniqueness in inverse AEM problems} 
\label{sec:identifying_and_addressing_non_uniqueness}

In the AEM literature DIMs are often evaluated, or compared, on an inverse problem without verifying whether the problem truly exhibits non-uniqueness.  One good reason for this may be that there is no general test for determining whether a problem exhibits non-uniqueness.  However, in general there is no guarantee that an inverse problem will be ill-posed, in which case DIMs may not offer any advantages over conventional regression models.  As we show, conventional models may achieve superior performance in these cases.  Furthermore, any performance differences between two DIMs on a well-posed problem cannot be caused by differences in their ability to address non-uniqueness, making such performance comparisons (between DIMs) potentially misleading.  

To mitigate these risks, we propose that a conventional NN always be employed as a baseline approach when solving inverse problems.  This ensures that a suitable model is included in case the problem is actually well-posed, and provides a baseline performance to ensure that there is some advantage to using DIMs.    Furthermore, one can treat the relative performance of a state-of-the-art DIM and conventional NN as a hypothesis test for the presence of non-uniqueness.  If the DIM achieves superior performance, it implies that its modeling assumption of non-uniqueness is (likely) more accurate for the problem than the NN's assumptions of uniqueness.  More precisely, we propose the following measure of non-uniqueness:
\begin{equation}
    \gamma =  \frac{r^{NA}}{r^{NN}}.
\end{equation}
Here $r^{NN}$ and $r^{NA}$ are the re-simulation errors of a conventional NN and the neural-adjoint DIM, when $T=1$.  We suggest the NN and the NA because they represent state-of-the-art deep conventional and inverse models, respectively. We use $T=1$ to avoid giving the NA an advantage by evaluating multiple of its solution proposals, while the NN can only submit one solution; in principle, if the problem is non-unique then the NA will yield superior performance even when $T=1$.  Finally, it is also important to ensure that the NA and NN are similarly-sized models (i.e., have similar number of free parameters), since model size can often impact performance.     


\begin{figure}[h!]
    \begin{center}
        \includegraphics[width=\textwidth]{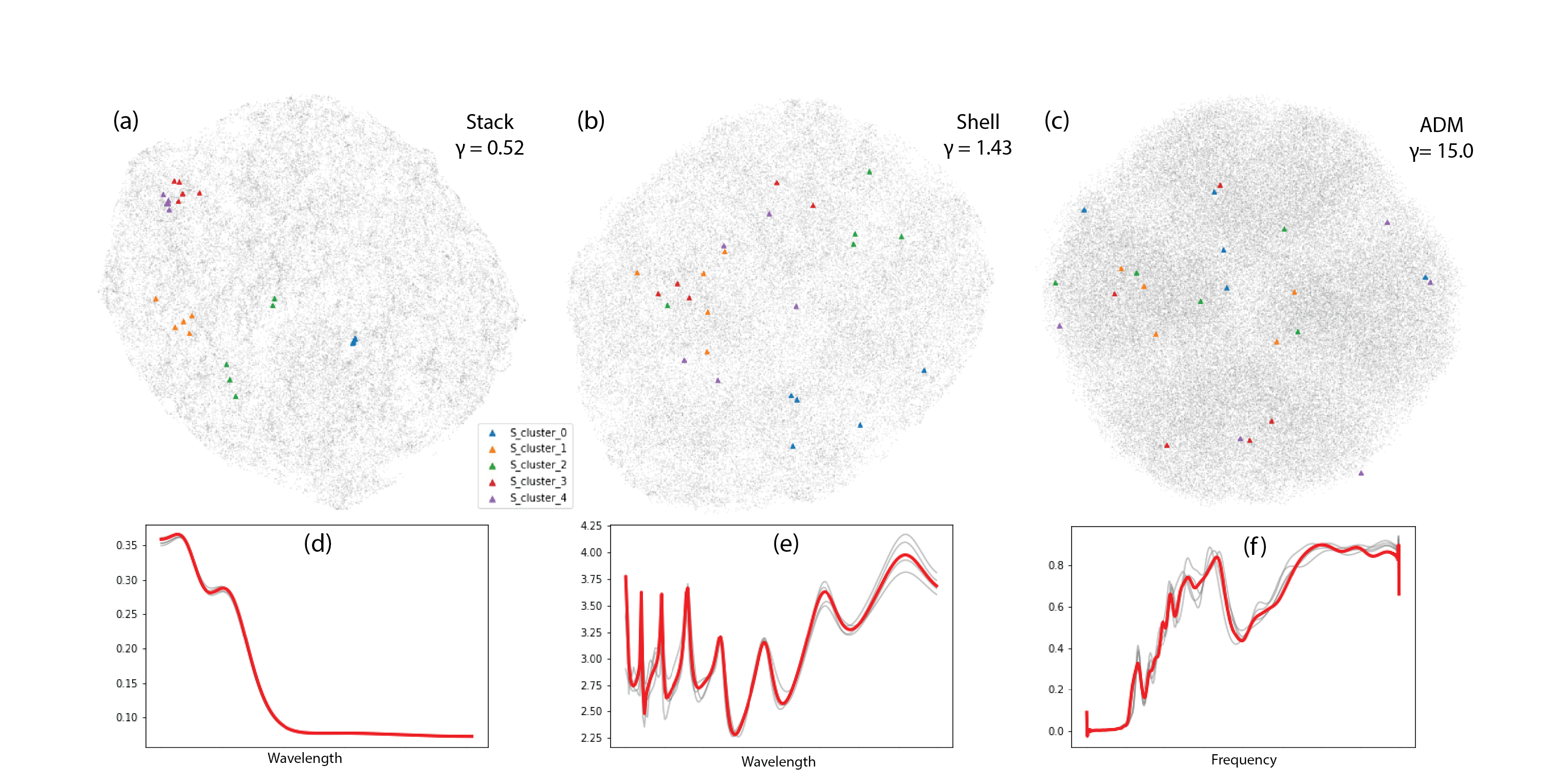} 
         \caption{Visualizing one-to-many of the benchmark datasets. (a-c) are the scatter plot of all geometry points in Stack, Shell and ADM in background in UMAP space (2d). On top of that, geometries that have similar spectra clusters (shown in (d-f)) are color-coded and plotted in triangles. There are 5 randomly selected clusters being plotted here. The more cluster the same color points are, the less one-to-many (in general) is the dataset.}
         \label{Fig:umap_scatter_plots_of_geometry_space}
     \end{center}
\end{figure}

\subsection{Non-uniqueness in the AEM benchmark tasks}

On top right corner of Fig \ref{Fig:umap_scatter_plots_of_geometry_space} we present the value of $\gamma$ for each of our three benchmark tasks.  The measure suggests that non-uniqueness is smallest in the Stack dataset, and largest in the ADM dataset.  In the Stack dataset the NN model achieves nearly \textit{twice} the accuracy of the NA method, which is the best-performing DIM at $T=1$ for this problem. This suggests that for some inverse problems, a simple regression model can indeed substantially outperform DIMs (when $T=1$), and therefore it can be costly exclude them when solving inverse problems.  Because the NN outperforms the DIMs, it suggests that non-uniqueness is not a major obstacle to solving this problem.  This is corroborated by the extremely low error of $6.34e-7$ achieved by the NN. From Fig. \ref{Fig:resimulation_error_versus_T}(e-g) this level of error is qualitatively low, providing a near-perfect match with the ground truth target spectra.  This also demonstrates that $\gamma$ is not simply a measure of problem difficulty either, because the overall accuracy of the inverse models ($r_{T}$) and $\gamma$ provide different rank-ordering of the benchmark tasks.   

Crucially, these results also imply that care must be taken when interpreting the performance differences between DIMs on the Stack problem.  Because there is relatively little non-uniqueness, differences in performance are largely due to factors that are unrelated to addressing non-uniqueness.  Although the Stack problem is well-posed, we do observe that as $T$ grows, the performance of the DIMs improves, and the DIMs do eventually outperform the conventional NN.  We hypothesize that this occurs because the DIMs are obtaining marginally more accurate approximations of the \textit{same} inverse solution found at $T=1$ by the NN, rather than identifying superior unique solutions. In the Appendix we provide further evidence that this is indeed the case.  


\subsection{Visualizing non-uniqueness in the AEM data}

To corroborate the $\gamma$ measure and its implications, we also present visual evidence of the non-uniqueness present in each benchmark task.  In Fig. \ref{Fig:umap_scatter_plots_of_geometry_space}(d-e) we randomly-sample a spectrum from each training dataset, $s_{1}$ (red), and then identify the four most similar spectra in the training dataset ($D_{tr}$), in terms of Euclidean distance (gray). We refer to this set of five total spectra as $S_{1}$.  In Fig. \ref{Fig:umap_scatter_plots_of_geometry_space}(a-c), we present a scatter plot of all designs in $D_{tr}$ along with the designs corresponding to $S_{1}$ shown in red.  We cannot directly scatter plot any of the designs because their dimensionality is greater than three, and therefore we use the UMAP \cite{umap} approach to reduce their dimensionality to two for visualization.   

For problems with a single unique solution, we expect that highly similar spectra will \textit{tend} to have designs that are also similar.  This is the case for the Stack problem in Fig. \ref{Fig:umap_scatter_plots_of_geometry_space}(a), while the points become increasingly distant for the Shell problem, and then again more distant for the ADM problem.  We repeat this process with four additional randomly selected spectra, $s_{i}, i\in {2,3,4,5}$ and plot the designs corresponding in each cluster with a different color.  The pattern of non-uniqueness among the three tasks holds for these additional spectra.  These visualizations provide additional evidence in support of the $\gamma$ measure.  We leave further theoretical analysis of $\gamma$ for future work.

\section{Conclusions} \label{sec:conclusion}

Recently deep inverse models have been found successful for solving inverse AEM problems (e.g., material design).  This has led to the rapid proliferation of different models for performing inverse design, but relatively little rigorous testing or comparison among them.  In this work we present the first benchmark performance comparisons of state-of-the-art deep inverse models (DIMs) on inverse AEM problems.  We selected three inverse AEM problems to include in the benchmark, which were chosen carefully to be relevant and diverse.  We then evaluated the performance of eight different DIMs on each of our three benchmark tasks.  Six of these DIMs were were employed in recent AEM studies, while two of them (the INN and cINN) are introduced to the AEM community for the first time in this study. 

Recent inverse AEM studies typically measure the accuracy of DIMs based upon the first solution that they propose for a target AEM scattering.  However, many recent DIMs can propose multiple solutions for a given target scattering and in this work we evaluated the performance of DIMs as a function of the number of solution proposals that they were permitted to make, denoted $T$.  Therefore, in our benchmark we evaluated DIM performance as a function of $T$, for $T\in [1,200]$   This measure is much richer than the conventional performance measure (equivalent to $T=1$), and we believe it better reflects how DIMs are often used in practice=.  

\textbf{Benchmarking results.} The results of our benchmark indicate that DIMs can achieve error levels ranging from nearly $10e-3$ for the Shell problem, to $10e-7$ on the Stack problem.  We believe this level of error (illustrated in Fig. \ref{Fig:resimulation_error_versus_T}(e-g)) is sufficient for a variety of applications, and demonstrates the overall effectiveness of DIMs for solving inverse AEM design problems.  We find that design quality improves substantially if we are allowed to evaluate multiple solutions (i.e., $T>>1$) from multi-solution DIMs (see Table \ref{Table:benchmark_inverse_model_characteristics}). Although the performance of DIMs varies by (i) task and (ii) the value $T$ being considered, we find that iterative methods such as the Neural-Adjoint and Genetic Algorithm tend to achieve the best performance. This performance advantage comes at greater computational cost during solution inference compared to other DIMs, however, these costs will usually be small compared to other computational costs.  

\textbf{Publication of resources for future benchmarking.} To make our experiments easily replicable, we publish code and documentation for our benchmark datasets and DIMs. Importantly, we also publish fast and easily-usable simulators, which are necessary to benchmark DIMs.  It is our hope that researchers will utilize these resources to rigorously evaluate new DIMs, as well as build upon our benchmark by adding new models and tasks.

\section*{Conflicts of interest}
Authors report no conflict of interest.

\section*{Acknowledgements}
We acknowledge funding from the Department of Energy under U.S. Department of Energy (DOE) (DESC0014372).


\bibliographystyle{unsrtnat}
\bibliography{references} 

\section{Appendix}

\section{Deep inverse models: further details}
Here we supply additional details of the deep inverse models being benchmarked in this work.

\subsection{Genetic Algorithm (GA).}

For GA, we provide extra details for how the mutation steps are carried out here: Parents are arranged into pairs, with each pair producing two child geometries via single-point crossover and point mutation. The probability of a pair undergoing crossover is defined by a hyper-parameter termed the crossover rate. In single-point crossover, a splice point is randomly selected along the geometry parameter vector with uniform probability. Parameters following the splice point are swapped between the pair of parents. The resulting population undergoes point mutation. In point mutation, parameters of each geometry in the population are replaced at random by new parameters in the domain with a probability determined by a hyper-parameter called the mutation rate. The resulting child population replaces the original population in the next iteration. The overall process can be found below:

\begin{figure}[h!]
\centering
 \includegraphics[width=0.5\textwidth]{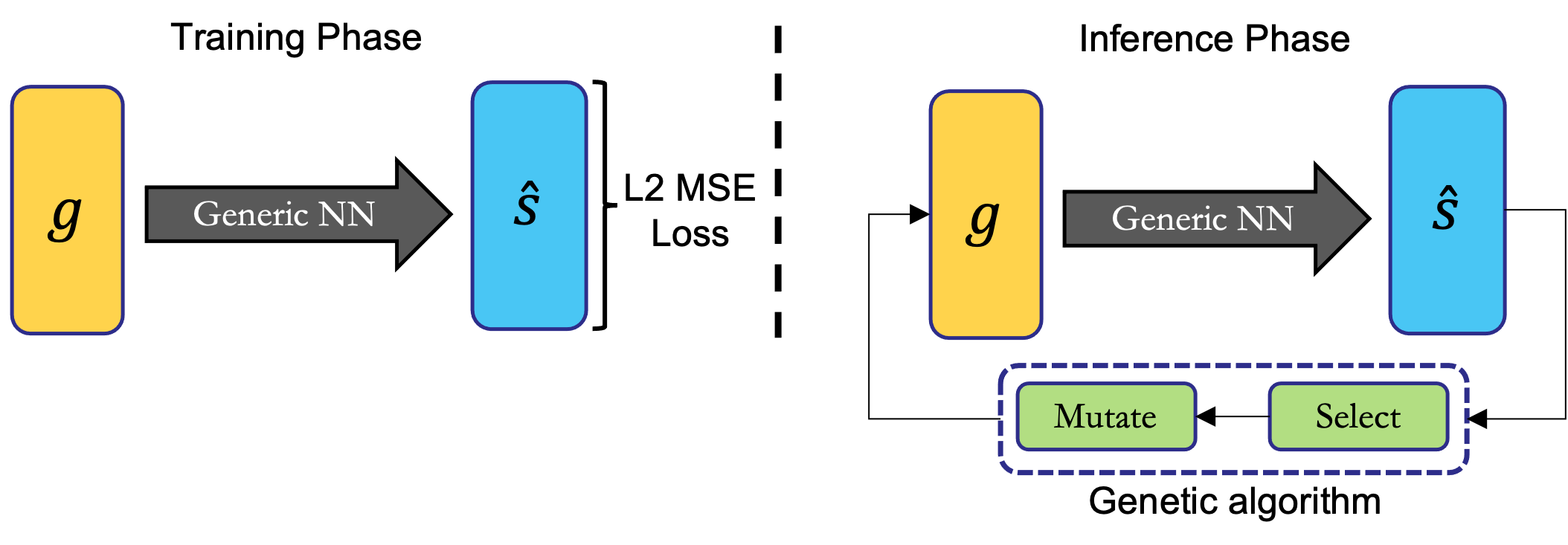}
 \caption{Schematic architecture of Genetic Algorithm}
 \label{arch_GA}
\end{figure}

\subsection{Neural Adjoint with Boundary Loss (NA).}
We supply the loss function and the process flow chart of the NA method. The loss function during inference time can be written as:
\begin{align}
    \mathcal{L}_{train} &= (\hat{f}(g) - s_{gt})^2 \\
    \mathcal{L}_{infer} &= (\hat{f}(\hat{g}) - g_{gt})^2 + \mathcal{L}_{bdy}\\
    \mathcal{L}_{bdy} &=  ReLU(|\hat{g} - \mu_g| - \frac{1}{2}R_g) \label{eq:na_grad}
\end{align}    
\begin{figure}[h!]
\centering
 \includegraphics[width=0.5\textwidth]{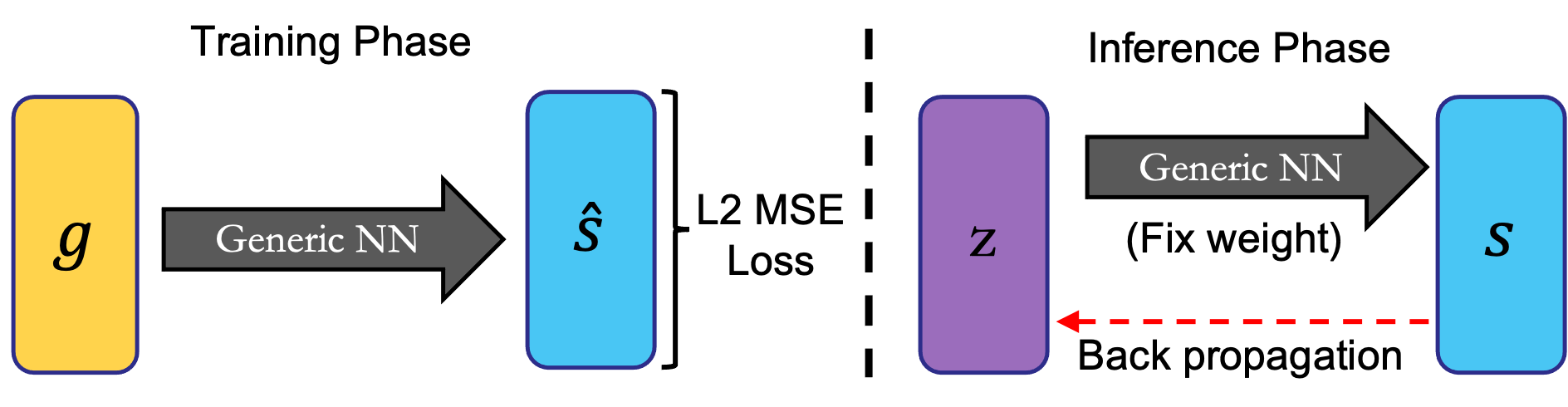}
 \caption{Schematic architecture of Neural Adjoint}
 \label{arch_NA}
\end{figure}
where $\hat{f}$ is the forward proxy function, $\mu_g$ and $R_g$ are the mean and range of the training set $g$. $\hat{g}$ is the trainable parameter of geometry during inference ($g_0$ randomly initialized and equivalent to $z$ in probabilistic methods) 
 
\subsection{Naive Neural Network (NN).}
We supply the loss function and the process flow chart of the NN method.
The loss function is as follows where $f^{-1}$ represents the naive neural network.
\begin{equation}
    \mathcal{L} = (f^{-1}(s) - g_{gt})^2
\end{equation}
\begin{figure}[h!]
\centering
 \includegraphics[width=0.5\textwidth]{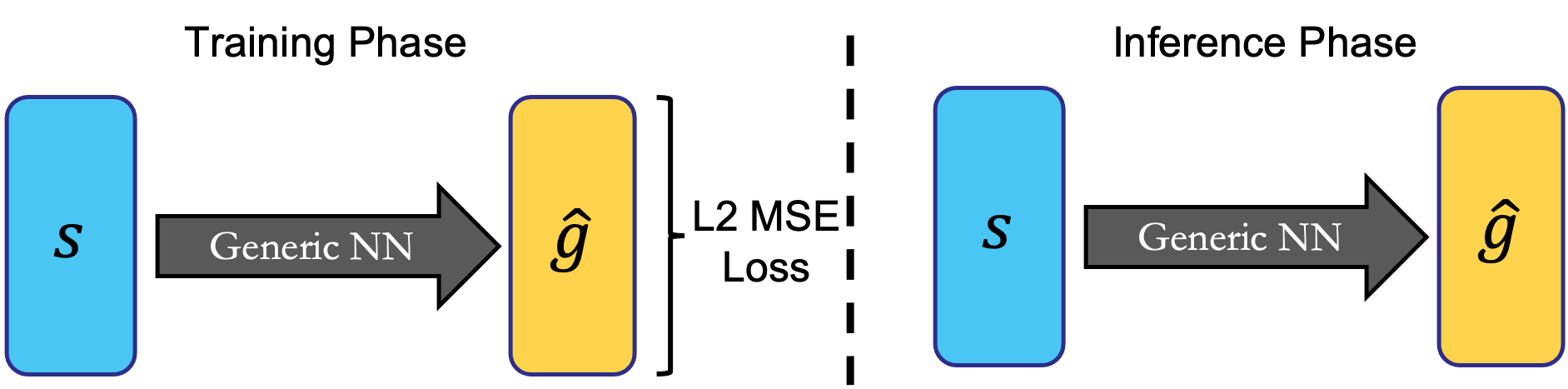}
 \caption{Schematic architecture of Naive Neural Network}
 \label{arch_NN}
\end{figure}

\subsection{Tandem Network (TD).}
We supply the loss function and the process flow chart of the TD method. The 2-stage training loss function for TD are as follows 
\begin{align}
    \mathcal{L}_1 &= (\hat{f}(g) - s_{gt})^2 \\
    \mathcal{L}_2 &= (\hat{f}(f^{-1}(s)) - g_{gt})^2 + \mathcal{L}_{bdy}
\end{align}
where the $\mathcal{L}_{bdy}$ is the same as defined in eq \ref{eq:na_grad}.
In our benchmarking tasks, to make a fair comparison, the same forward network $\hat{f}$ is used for TD, the Genetic Algorithm, and the Neural Adjoint with boundary loss.
\begin{figure}[h!]
\centering
 \includegraphics[width=0.5\textwidth]{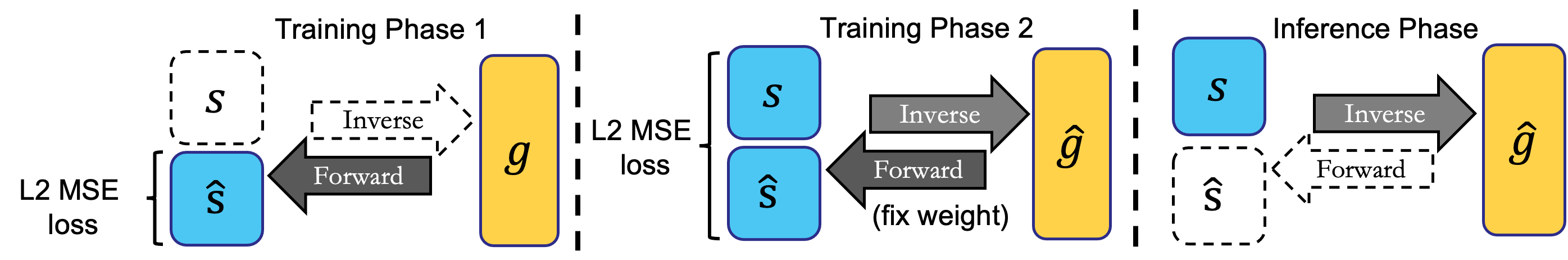}
 \caption{Schematic architecture of Tandem model}
 \label{arch_TD}
\end{figure}

\subsection{Mixture Density Network (MDN).}
We supply the loss function and the process flow chart of the MDN method. The loss function is as follows where $p_i, \mu_i, \Sigma_i$ represents the probability of geometry coming from Gaussian distribution i and the mean and variance of Gaussian distribution i.
\begin{equation}
    \mathcal{L} = - \log(  \sum_i p_i *|\Sigma_i^{-1}|^{\dfrac{1}{2}} * \exp(-\dfrac{1}{2} (\mu_i - g)^{T} \Sigma_i^{-1} (\mu_i - g)))
\end{equation}
\begin{figure}[h!]
\centering
 \includegraphics[width=0.5\textwidth]{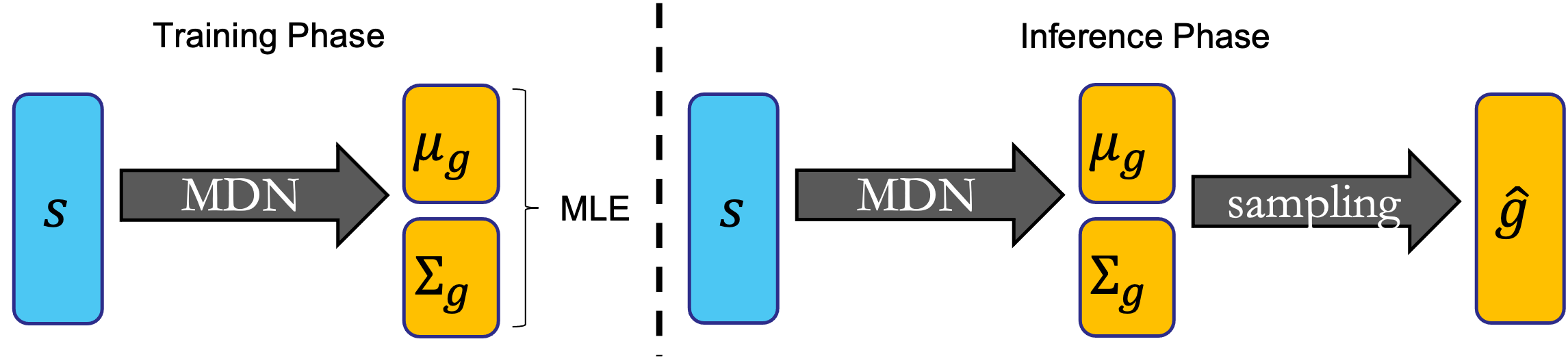}
 \caption{Schematic architecture of Mixture Density Network}
 \label{arch_MDN}
\end{figure}

\subsection{Conditional Variational Autoencoder (VAE).}
We supply the loss function and the process flow chart of the TD method.  The following loss term captures the training process of a VAE. $z$ is forced to follow a normal distribution ($\mu_z, \sigma_z$ is the mean and variance of $z$) and trained by minimizing the KL divergence between the encoder $q(z|g,s)$ and a normal distribution. A mean squared reconstruction loss encourages the decoder $p(g|z,s)$ to accurately reconstruct the input geometry given a latent vector sampled from $z$ and the condition $s$. $\alpha$ is a hyper-parameter trading off the reconstruction and normality of the latent space and needs to be tuned carefully on a validation set.
\begin{equation}
    \mathcal{L} = (g - \hat{g})^2 - \frac{\alpha}{2} \cdot (1 + log\sigma_z + \mu_z^2 - \sigma_z) \label{ELBO}
\end{equation}
\begin{figure}[h!]
\centering
 \includegraphics[width=0.5\textwidth]{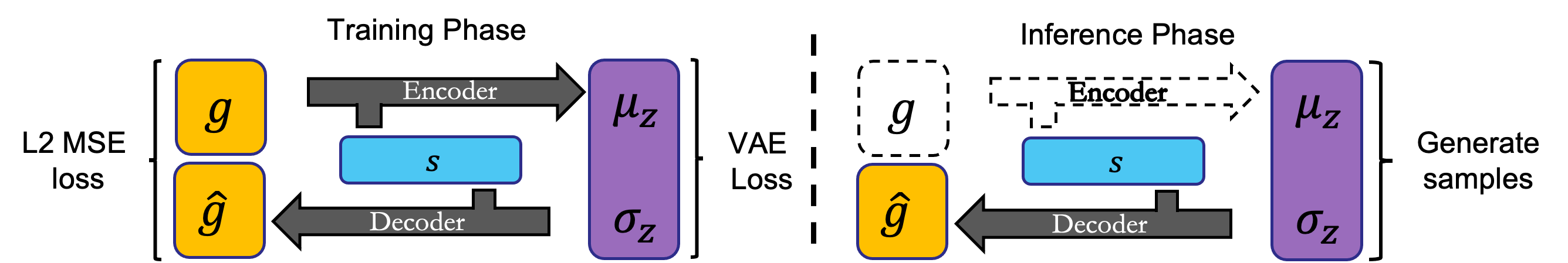}
 \caption{Schematic architecture of Conditional Variational Autoencoder}
 \label{arch_VAE}
\end{figure}

\section{Forward surrogate models}

\subsection{Forward model performance compared to original work}
Our forward models are optimized for each benchmarking task. For the Shell problem we achieved a forward model MRE of 0.65\%, compared to 1.5\% in the original work over 5,000 test samples. For the Graphene-Si\textsubscript{3}N\textsubscript{4} 2D Multi-layer Stack, our NN achived MRE of 0.4\% while Chen et al. reported with MRE of 3.2\% given 1,000 samples\cite{Chen2019}. On the Super-unit Cell benchmark solved by Deng et al \cite{deng2021}, we get similar MSE values (1.2e-3) on $T=1$ performance.
\section{Model details}
\subsection{Model size}
Here we present the table of model size in number of trainable parameters in Table \ref{Table:total_trainable_parameters}.

\begin{table}[h!]
    \centering
    \begin{tabular}{cccc}
        \toprule
        Model                                       & Multi-layer Stack (Chen)     &  Multi-layer Shell (Peurifoy)   &  Super-unit Cell (Deng)  \\
        \midrule
        NN       & 16M &  25M & 36M   \\
        TD                                 & 3M &  93M & 47M         \\
        GA             & 4M &   40M & 17M \\
        NA                   & 4M &   40M & 17M \\
        VAE       & 12M&   18M & 14M \\
        INN       & 8M&   35M & 25M \\
        cINN & 3M &   35M & 23M\\
        MDN               & 0.4M &   0.1M & 1M \\
        \bottomrule
    \end{tabular}
\caption{Model size reflected by the total number of trainable parameters. Note that the tandem model is typically bigger due to it make use of a forward and a backward model and MDN models are small due to after hyper-parameter sweep all larger models exhibit much worse performance than the current selected best one.}
\label{Table:total_trainable_parameters}
\end{table}

\section{Stack dataset DIM performance improvement over T}
\begin{figure}[h!]
\centering
 \includegraphics[width=\textwidth]{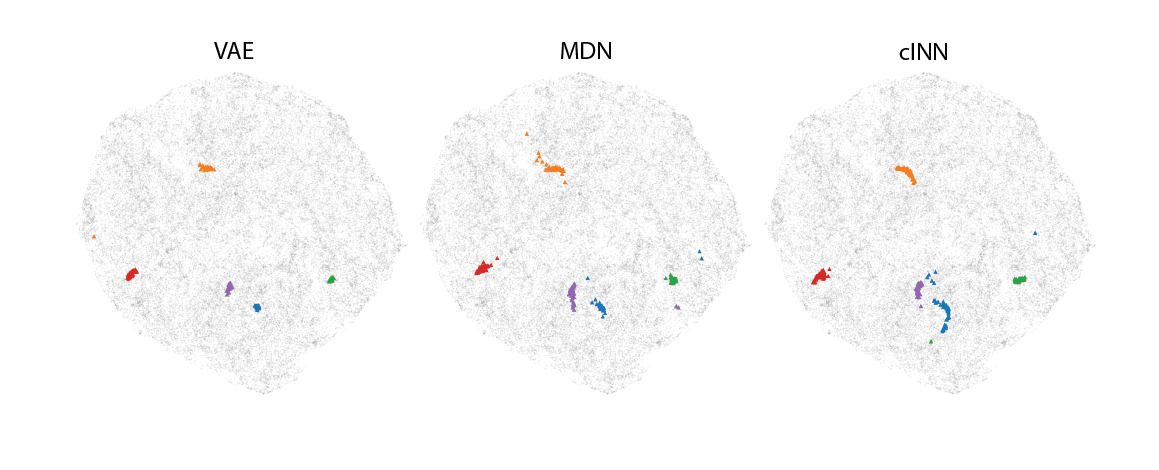}
 \caption{Cluster effect of retrieved $\hat{g}$ of Stack dataset top-3 performing algorithms (at T=200) top 50 proposed solution. Same as the main text umap plot, the background grey points are the scatter point of all point of Stack dataset. On top of that, for each of the 5 random spectra (labelled by different color), 200 random $\hat{g}$ was retrieved by individual models and top 50 chosen by the real simulator. }
 \label{fig:app_stack_exploration}
\end{figure}

In main text, we discussed that the performance improvement over T for DIMs on Stack dataset is mainly due to "jittering" around the solution space instead of actually finding new solutions, here we show empirical evidence Fig \ref{fig:app_stack_exploration} supporting the claim.

As Fig \ref{fig:app_stack_exploration} shows, all top 50 points cluster nicely in local regions, supporting the statement that these DIMs only explore locally for the Stack dataset (with the lowest non-uniqueness) when T gets larger.

\section{Another metric for one-to-many based on neighbourhood}
We also provide some quantified measurement of the distance that is not distorted by the dimensionality reduction, we provide $D_r$ (defined below) to reflect the "localness" of the clustered points, normalized by the pairwise distance of the whole dataset. $D_r \approx 1$ would mean that the cluster is closer to randomly drawn from the whole dataset, therefore a localized draw would have a $D_r < 1$.
\begin{align*}
     D_r = & \dfrac{\text{Avg pairwise $G$ distance within cluster}}{\text{Avg pairwise $G$ distance between all points}}\\
     = & \dfrac{\dfrac{1}{C}\sum_c^C \dfrac{1}{K(K-1)} \sum_i^K \sum_{j\neq i}^K |G_{c,i} - G_{c,j}|^2}{\dfrac{1}{N(N-1)} \sum_i^N \sum_{j\neq i}^N |G_i - G_j|^2  }
\end{align*}
where $C$ is the number of cluster (5 in our case), $K$ is the number of data points in each cluster (5 in our case), $N$ is the size of the whole dataset, $G_{c,i}$ represents the i-th geometry of cluster $c$. 

The $D_r$ value is labelled on the right top corner of Fig \ref{Fig:umap_scatter_plots_of_geometry_space}. As this "closeness" measure increase from order: Stack < Shell < ADM, we see the same trend quantitatively with the perceived one-to-manyness on the plot as well.

\begin{table}[h]
    \begin{center}
    \begin{tabular}{cccc}
        \toprule
        Dataset      &      Stack       &       Shell    &       ADM \\
        \midrule
        $\gamma$  &        0.52        &        1.43     &      15.0  \\
        $D_r$       &       0.24         &         0.58     &     0.84 \\
        \bottomrule
  \end{tabular}
  \caption{Ratio of forward problem MSE and inverse problem MSE and perceived one-to-manyness}
  \label{Table:non_uniqueness_measures}
  \end{center}
\end{table}

\section{Hardware for training used}
Here we list the hardwares we used in the model training and inference to provide context for our running time. We are using NVIDIA GTX3090 cards for GPU, AMD 3990X (64 cores, 2.90GHz, 256 MB cache), RAM: 256GB.

\end{document}